\pdfoutput=1

\documentclass[11pt]{article}

\usepackage[preprint]{acl}
\usepackage{float} 
\usepackage{amsmath}
\usepackage{amsfonts}
\usepackage{times}
\usepackage{latexsym}
\usepackage{bbm}

\let\cite\citep

\usepackage[textsize=scriptsize]{todonotes}
\usepackage{xspace}

\makeatletter
\newcommand*\iftodonotes{\if@todonotes@disabled\expandafter\@secondoftwo\else\expandafter\@firstoftwo\fi}  %
\makeatother

\definecolor{darkblack}{rgb}{0.0,0.0,0.5}
\definecolor{darkgreen}{rgb}{0.0, 0.42, 0.24}
\definecolor{lightgreen}{rgb}{0.52, 0.73, 0.4}
\definecolor{darkgray}{rgb}{0.4,0.4,0.4}
\definecolor{darkblue}{rgb}{0.0,0.0,0.5}
\definecolor{darkpurple}{rgb}{0.5,0.2,0.8}
\definecolor{lightpurple}{rgb}{0.8,0.5,1}

\usepackage{cleveref}
\crefname{figure}{Figure}{Figures}
\crefname{table}{Table}{Tables}
\crefname{appendix}{Appendix}{Apps.}
\crefname{section}{\S}{\S\S}
\crefformat{section}{\S#2#1#3} %
\crefname{equation}{Eq.}{Eqs.}
\crefname{algorithm}{Alg.}{Algs.}
\crefname{algocf}{Alg.}{Algs.}

\usepackage{wrapfig}
\usepackage{tabularx}
\usepackage{booktabs}
\usepackage{dsfont}
\usepackage{multirow}
\usepackage{array}

\usepackage{caption}
\usepackage{listings}
\usepackage{adjustbox}

\usepackage[T1]{fontenc}

\usepackage[utf8]{inputenc}

\usepackage{microtype}

\usepackage{inconsolata}

\usepackage{graphicx}

\title{Probing for Arithmetic Errors in Language Models}

\author{%
  Yucheng Sun\textsuperscript{*}  \quad Alessandro Stolfo\thanks{\hspace{0.2mm}Equal contribution.} \quad Mrinmaya Sachan \\
  ETH Z\"urich \\ 
  \texttt{\{yucsun, stolfoa\}@ethz.ch}}

\begin{document}
\maketitle
\begin{abstract}
We investigate whether internal activations in language models can be used to detect arithmetic errors. Starting with a controlled setting of 3-digit addition, we show that simple probes can accurately decode both the model’s predicted output and the correct answer from hidden states, regardless of whether the model’s output is correct. Building on this, we train lightweight error detectors that predict model correctness with over 90\% accuracy. 
We then extend our analysis to structured chain-of-thought traces on addition-only GSM8K problems and find that probes trained on simple arithmetic generalize well to this more complex setting, revealing consistent internal representations. Finally, we demonstrate that these probes can guide selective re-prompting of erroneous reasoning steps, improving task accuracy with minimal disruption to correct outputs. Our findings suggest that arithmetic errors can be anticipated from internal activations alone, and that simple probes offer a viable path toward lightweight model self-correction.
\end{abstract}

\section{Introduction}

Large language models have recently shown strong performance on mathematical problem solving and reasoning \cite[\textit{inter alia}]{openai2024openaio1card, geminiteam2024gemini15unlockingmultimodal, shao2024deepseekmathpushinglimitsmathematical,deepseekai2025deepseekr1incentivizingreasoningcapability}, making mathematical ability an increasingly central axis for benchmarking model capabilities \cite{glazer2024frontiermathbenchmarkevaluatingadvanced}. This surge in capability has sparked growing interest in understanding how these models internally process numerical information and execute arithmetic reasoning \cite{nanda2023progress, stolfo-etal-2023-mechanistic, hanna2023how, zhong2023the, zhou2024pretrained, nikankin2025arithmetic, lindsey2025biology}.

In particular, recent studies have investigated how pre-trained language models represent numerical quantities \cite{levy-geva-2025-language, kantamneni2025language, zhu-etal-2025-language}. However, despite increasing insight into the structure of these representations, their practical use for improving model behavior remains limited. At the same time, analyzing hidden representations has proven useful for identifying model failures and hallucinations in other domains \cite{kadavath2022languagemodelsmostlyknow, azaria-mitchell-2023-internal, yuksekgonul2024attention, chen2024inside, orgad2025llms}.

\begin{figure}[t]
    \centering
    \includegraphics[width=0.42\textwidth]{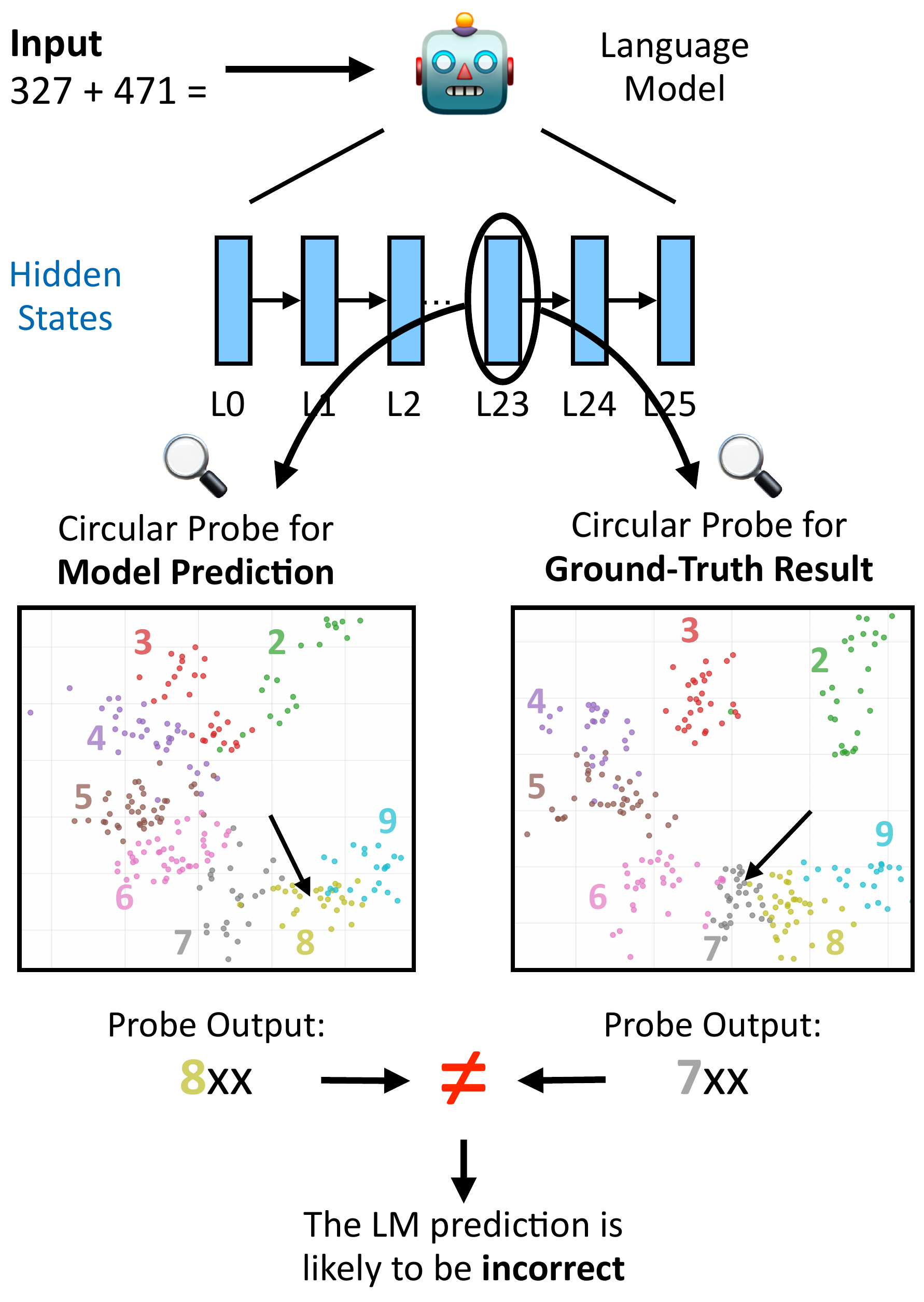}
    \caption{\textbf{Detecting Arithmetic Errors from Hidden States.}
    We investigate whether internal activations in a language model reveal when its arithmetic predictions are incorrect. We train simple probes to decode both the model’s output and the correct answer. The probes' output can serve as a reliable signal of model error.%
    \vspace{-3mm}
    }
    \label{fig:fig1}
\end{figure}

Motivated by these findings, we investigate whether internal model activations can be leveraged to detect arithmetic errors in language models. We begin with a simple setting: the model is prompted with 3-digit addition queries (e.g., ``327+471''), and we analyze its hidden activations using simple probes inspired by representational and circuit-level findings. We show that not only it is possible to recover the language model’s prediction, but also to predict the ground-truth result from the model’s hidden states, independently of whether the model’s output is correct. Motivated by this observation, we adapt the probing method to predict whether the model’s final answer will be correct  (\cref{fig:fig1}), achieving over 90\% accuracy in predicting model correctness on a balanced dataset.

We then extend this analysis to a more complex setting, where the model is asked to solve math word problems only requiring addition \cite{cobbe2021trainingverifierssolvemath} using a structured chain-of-thought (CoT) format \cite{wei2022chain}, in which intermediate steps are expressed as equations (e.g., \texttt{<a+b=c>}).
Remarkably, we find that the same probes trained on simple arithmetic queries can be applied directly to this setting, maintaining over 80\% accuracy in detecting whether the model is producing correct intermediate results.

Finally, we explore the practical benefits of these probes by using them as weak oracles to identify potentially erroneous steps within the model’s reasoning traces. We design a re-prompting mechanism that selectively revisits flagged steps, leading to an overall correction of up to 11\% of the wrong reasoning steps, without compromising any of the correct ones.
These results suggest that internal representations of numerical quantities are robust across contexts, and can be exploited to detect and correct model errors with lightweight probing methods.\looseness=-1

\section{Probing for Numerical Representations}
\label{sec:probing_intro}

We aim to understand how information about arithmetic operations is internally represented in a language model. To this end, we consider a controlled setting in which the model is prompted with simple 3-digit addition queries and analyze its hidden activations at the point just before it produces an output. Specifically, we ask whether numerical quantities are encoded in a form that can be recovered by lightweight probes. We begin with a representation analysis (\cref{sec:rep_analysis}), then introduce a set of probing methods to decode this information (\cref{sec:probes}).

\subsection{Representation Analysis}
\label{sec:rep_analysis}
Prior work has shown that language models represent periodic concepts such as the days of the week in circular forms \cite{engels2025not}. In the mathematical domain, similar findings suggest that numerical quantities can be encoded in circular \cite{levy-geva-2025-language}, linear \cite{zhu-etal-2025-language}, or hybrid helix-like \cite{kantamneni2025language} geometries. 
To better understand how numerical information is encoded in our setting, we conduct a representation analysis on the instruction-tuned version of Gemma 2 2B \cite{gemmateam2024gemma2improvingopen}, focusing on its behavior when solving 3-digit addition problems (e.g., ``327+471'').\footnote{The Gemma tokenizer splits numbers into individual digits. For our experiments, we focus on predicting the first digit of the result (i.e., the hundreds place). This choice captures the majority of the model’s errors in this setting (evidence for this in \cref{app:error_analysis}). Results for a model with a different tokenization scheme (Phi-3), which confirm our findings, are reported in \cref{app:phi3}.}

We synthetically generate arithmetic queries by sampling 800 pairs of operands $(a, b) \in \{100,\dots,999\}^2$, such that $a+b < 1000$. Then, we feed the model the prompt ``$a$+$b$='', and examine the model’s internal activations at the position of the equals sign (``=''). This token position immediately precedes the model’s output and is expected to carry information about the computed result \cite{stolfo-etal-2023-mechanistic, nikankin2025arithmetic}.

We apply principal component analysis (PCA) to the residual stream activations at this position across all layers of the model, highlighting in different colors the hidden states associated with different hundreds digits in the ground-truth result. We observe that as depth increases, the representations of individual digits become increasingly structured and separable. In particular, deeper layers exhibit clearer clustering by digit number, with the top principal components revealing a circular layout similar to those observed in prior studies.

\begin{figure}[t]
    \centering
    \includegraphics[width=0.41\textwidth]{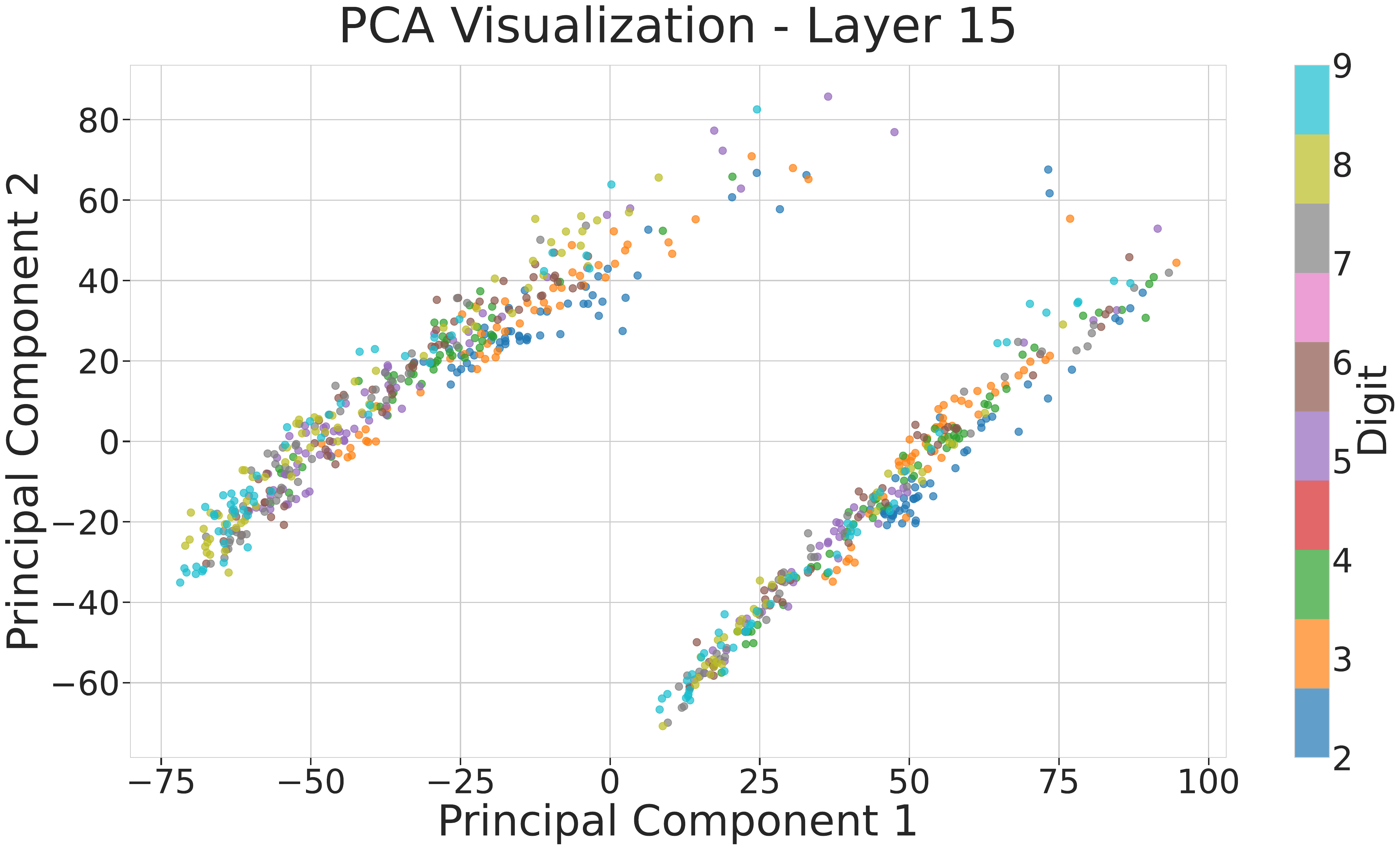}
    
    \vspace{0.45em}
    \includegraphics[width=0.41\textwidth]{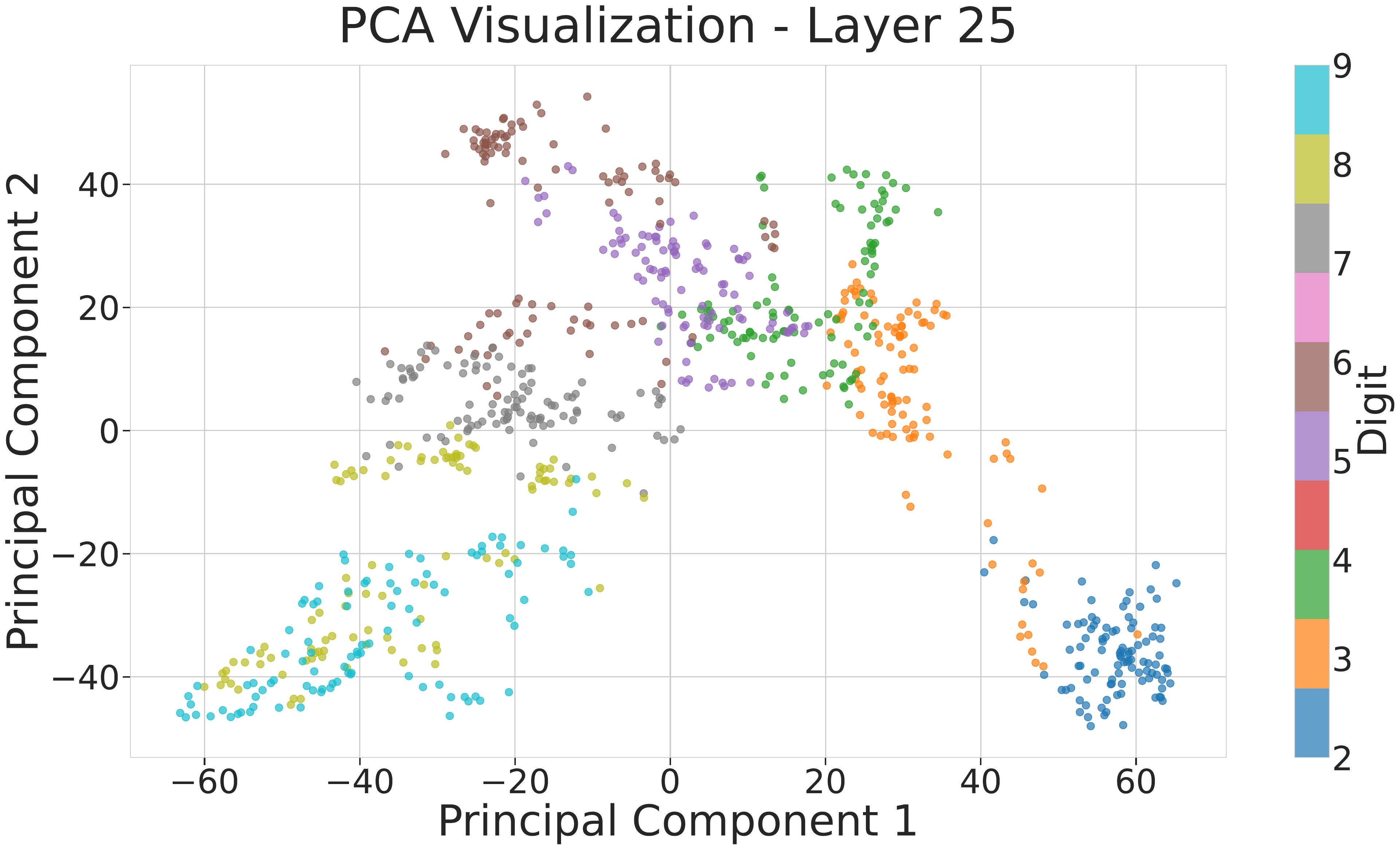}
    \caption{\textbf{PCA of Residual Stream Activations.} PC projections colored by the hundreds digit of the ground-truth result. Representations become more structured with depth, showing clear digit clusters and a circular layout in deeper layers.}
    \label{fig:pca}
\end{figure}

To illustrate this progression, we selected two representative layers for visualization. As shown in \cref{fig:pca}, layer 15 shows no clear digit separation, whereas layer 25 exhibits both stronger clustering and a visible circular layout. This progression supports the hypothesis that the model gradually builds a more geometric and abstract encoding of numerical concepts over its depth, and is consistent with what previous work observed.

\subsection{Probing Methods}
\label{sec:probes}

To quantitatively assess how numerical information is encoded in the model’s internal representations, we design a set of lightweight probing methods. Building on the insights from our representational analysis, which indicated a circular structure in digit encodings at deeper layers, we treat circular probing as a natural baseline. However, we also explore alternative probing approaches to evaluate different ways of extracting numerical information from the same representations.

As in the previous section, we focus on the residual stream activations at the equals sign. Each probe is trained and evaluated independently at each layer of the model, allowing us to track how the accessibility of digit-level information evolves across the model’s depth.
We describe the implementation of each probe below.

\paragraph{Circular Probe.}
Denote the residual-stream hidden states of a language model at layer $l$ by $\mathbf{x}_l \in \mathbb{R}^{d_\mathrm{model}}$.\footnote{By ``residual stream,'' we indicate per-token hidden state with dimensionality $d_\mathrm{model}$ consisting of the sum of all previous component outputs \cite{elhage2021mathematical}.} Let $y \in \{0, \dots, 9\}$ be the label that the probe is tasked to predict.
We design a circular probe that projects the model's hidden states onto a plane, then uses the angle that such projection point forms with the origin as its output. 
More formally, let the probe weights be two vectors $\mathbf{w}_1, \mathbf{w}_2 \in \mathbb{R}^{d_\mathrm{model}}$. We compute the probe's output $\hat{y}$ as
\begin{align}
    \theta =& \mathrm{atan2} \big(\mathbf{w}_1^\top\mathbf{x}_l, \mathbf{w}_2^\top\mathbf{x}_l \big) \in [0, 2\pi),\\
    \hat{y} =&\ \theta \cdot \frac{10}{2\pi}.
\end{align}

The probe is trained using smooth $\ell_1$ loss \cite{girshick2015fastrcnn} between the prediction $\hat{y}$ and the label $y$.
The probe is optimized using the AdamW optimizer \cite{loshchilov2018decoupled}.

\paragraph{Linear Probe.}
Motivated by prior work showing that numerical quantities can be linearly decoded from language model activations \cite{zhu-etal-2025-language, kantamneni2025language}, we test whether any arithmetic information in our setting can be recovered by a simple linear probe:
\begin{equation}
    \hat{y} = \mathbf{w}^\top \mathbf{x}_l + b,
\end{equation}
where $\mathbf{w} \in \mathbb{R}^{d_\mathrm{model}}$ and $b \in \mathbb{R}$ are the probe's parameters. %
The probe is trained using $\ell_2$ regularization.\footnote{Following \citet{zhu-etal-2025-language}, we use regularization weight $\lambda=0.1$.}
At inference time, the probe’s output is rounded to the nearest integer to produce a discrete prediction.

\paragraph{Multi-Class Logistic Regression.}
We also consider a variant of the linear probe that treats digit prediction as a multi-class classification problem.
Specifically, we associate a distinct weight vector $\mathbf{w}_i$ with each digit $i \in \{0, \dots, 9\}$, and compute the probe’s output as the digit corresponding to the maximum logit:
\begin{equation}
    \hat{y} = \arg\max_{i} \left( \mathbf{w}_i^\top \mathbf{x}_l \right).
\end{equation}
The probe is trained using cross-entropy loss and optimized with the Adam optimizer \cite{kingma2017adammethodstochasticoptimization}.

\begin{figure*}[t]
    \centering
   \includegraphics[width=0.92\textwidth]{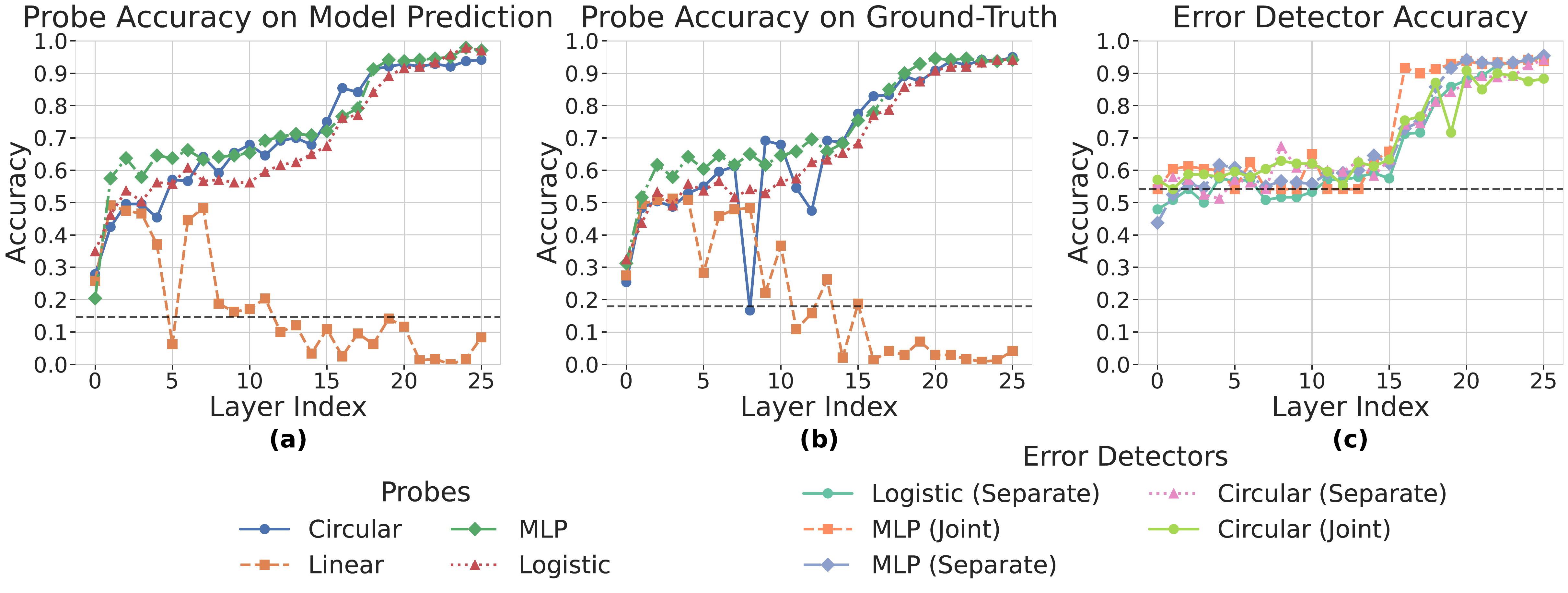}
    \caption{\textbf{Probing 3-Digit Arithmetic Queries.}
(a) Probes recover the model’s output with high accuracy in deeper layers; linear probes perform poorly.
(b) Ground-truth digits are similarly decodable, suggesting correct answers are often internally represented.
(c) Error detectors show that model correctness can be inferred from hidden states.
The dashed lines indicate the accuracy of the majority class baseline.}
    \label{fig:pure_setting_probes}
\end{figure*}

\paragraph{MLP Probe.}
Finally, we experiment with a more expressive probe based on a multi-layer perceptron (MLP). The MLP consists of a single hidden layer with ReLU activation and a hidden dimensionality of 512. As in the logistic probe, the output layer produces 10 logits corresponding to the digit $i \in \{0, \dots, 9 \}$. The prediction is given by:
\begin{equation}\nonumber
    \hat{y} = \arg\max_i \bigg(\mathbf{W}_2^\top \ \mathrm{ReLU}\Big(\mathbf{W}_1^\top \mathbf{x}_l + \mathbf{b}_1 \Big) + \mathbf{b}_2\bigg),
\end{equation}
where $\mathbf{W}_1 \in \mathbb{R}^{d_\mathrm{model} \times 512}, \mathbf{W}_2 \in \mathbb{R}^{512 \times 10}, \mathbf{b}_1 \in \mathbb{R}^{512}$, and $\mathbf{b}_2 \in \mathbb{R}^{10}$ are the probe's parameters. This probe  is trained using cross-entropy loss and optimized with Adam.

\section{Probing 3-Digit Arithmetic Queries}
\label{sec:probe_arith}

We begin our analysis in the arithmetic setting described in \cref{sec:probing_intro}, where the model is prompted to solve three-digit addition problems (e.g., ``652+185''). We use a few-shot prompt containing two exemplars of addition (data details are provided in \cref{app:pure_data}).
We organize our analysis in three stages: first, we test whether the probes can recover the model’s predicted output (\cref{sec:pure_output}); second, we examine whether they can recover the ground-truth result (\cref{sec:pure_ground_truth}); and third, we test whether combining the two signals allows us to predict whether the model is correct on a given query (\cref{sec:pure_correctness}).

\subsection{Can Probes Predict the Model’s Output?}
\label{sec:pure_output}
In this first experiment, we train probes to predict the first digit of the model’s output, i.e., the digit that the model intends to produce as the result of the addition. We generate a dataset of queries of the form ``$a$+$b$'', where $a$ and $b$ are three-digit integers. For each query, we record the model’s prediction and evaluate whether it matches the correct result. We construct a  dataset with 800 queries, balanced across both output digits and model correctness, and split it into 70\% training and 30\% evaluation sets. Each probe is trained for 10,000 epochs. We apply and evaluate each probe independently across all 26 layers of Gemma 2B IT.%

The results are shown in \cref{fig:pure_setting_probes}a. The linear probe fails to recover meaningful information from the model’s hidden states, maintaining low accuracy across all layers. In contrast, the circular, logistic, and MLP probes achieve significantly higher performance, with accuracy progressively improving across layers and plateauing around 92\% in the final 4-5 layers.

Two key observations emerge. First, the accuracy gains across layers align with prior work showing that the final MLP blocks in transformer models are responsible for computing arithmetic outputs \cite{stolfo-etal-2023-mechanistic, nikankin2025arithmetic}. Second, despite its simplicity and small number of parameters, the circular probe performs on par with both the MLP and logistic probes. This supports the hypothesis--grounded in our representational analysis (\cref{sec:probing_intro})--that digit information is encoded in a circular manner, and that a probe matched to this geometry is sufficient to extract it.

\subsection{Can Probes Recover the Ground-Truth Answer?}
\label{sec:pure_ground_truth}

We next test whether probes can recover the correct answer to an arithmetic query, even when the model’s own prediction is incorrect. Concretely, we train probes to predict the first digit of the ground-truth sum, using the same dataset and training procedure as in \cref{sec:pure_output}. The only difference is that the probe labels now correspond to the true answer rather than the model’s prediction.

Accuracy results across all layers are shown in \cref{fig:pure_setting_probes}b. Surprisingly, the trends follow those observed in the previous experiment: accuracy increases steadily with layer depth, and the MLP, logistic, and circular probes all reach performance levels comparable to those observed when predicting the model’s output (>90\%). This suggests that the correct result is encoded in the model’s internal representations and is accessible to simple probes, even when the model ultimately generates an incorrect answer. This raises a natural question: are probes (1) genuinely recovering a representation of the correct answer from the model’s hidden states (despite the model’s failure to output it), or are they (2) simply learning to perform the arithmetic themselves during training, by extracting operand information and learning to compute the result?

To disentangle these hypotheses, we conduct a follow-up experiment. We train MLP probes at an early layer (layer 5) to decode the input operands, both as complete numbers and as individual digits. If hypothesis (2) were correct (i.e., the probes were solving the arithmetic task themselves), then we would expect their performance on ground-truth prediction to match their performance on operand prediction. In other words, if the probes have access to the operands and are capable of computing the result, they should be able to predict the correct answer whenever the operand representations are available.

The results, summarized in \cref{tab:operand_digits}, show that operand information is indeed recoverable at early layers, with probes achieving nearly perfect accuracy. However, as observed in \cref{fig:pure_setting_probes}, the probes can only predict the ground-truth result with high accuracy in deeper layers. 
This discrepancy suggests that the probes are not simply performing the arithmetic task themselves, lending support to hypothesis (1) over (2). At the same time, the evidence does not fully justify the stronger claim that probes recover a clean representation of the correct answer. Instead, a more plausible interpretation is that probes learn to refine partial, noisy information about the result that is already present in the residual stream, and that the model itself may fail to fully decode through its output head. %
This interpretation is consistent with recent findings showing that language models often encode correct answers internally even when their output is incorrect \cite{orgad2025llms, gekhman2025insideouthiddenfactualknowledge}.

\begin{table}[t]\small
\centering
\begin{tabular}{lc}
\toprule
\textbf{Operand Component} & \textbf{Accuracy} \\
\midrule
Operand \#1 - Hundreds & 0.9500 \\
Operand \#1 - Tens & 0.8833 \\
Operand \#1 - Ones & 0.9500 \\
Operand \#2 - Hundreds & 0.9833 \\
Operand \#2 - Tens & 0.9958 \\
Operand \#2 - Ones & 1.0000 \\
\bottomrule
\end{tabular}
\caption{\textbf{Operand Decoding Accuracy at Layer 5.}
MLP probes can reliably recover operand values from early-layer activations, indicating that input numbers are explicitly represented in the model’s residual stream.}
\label{tab:operand_digits}
\end{table}

\begin{figure*}[t]
    \centering
    \includegraphics[width=0.92\textwidth]{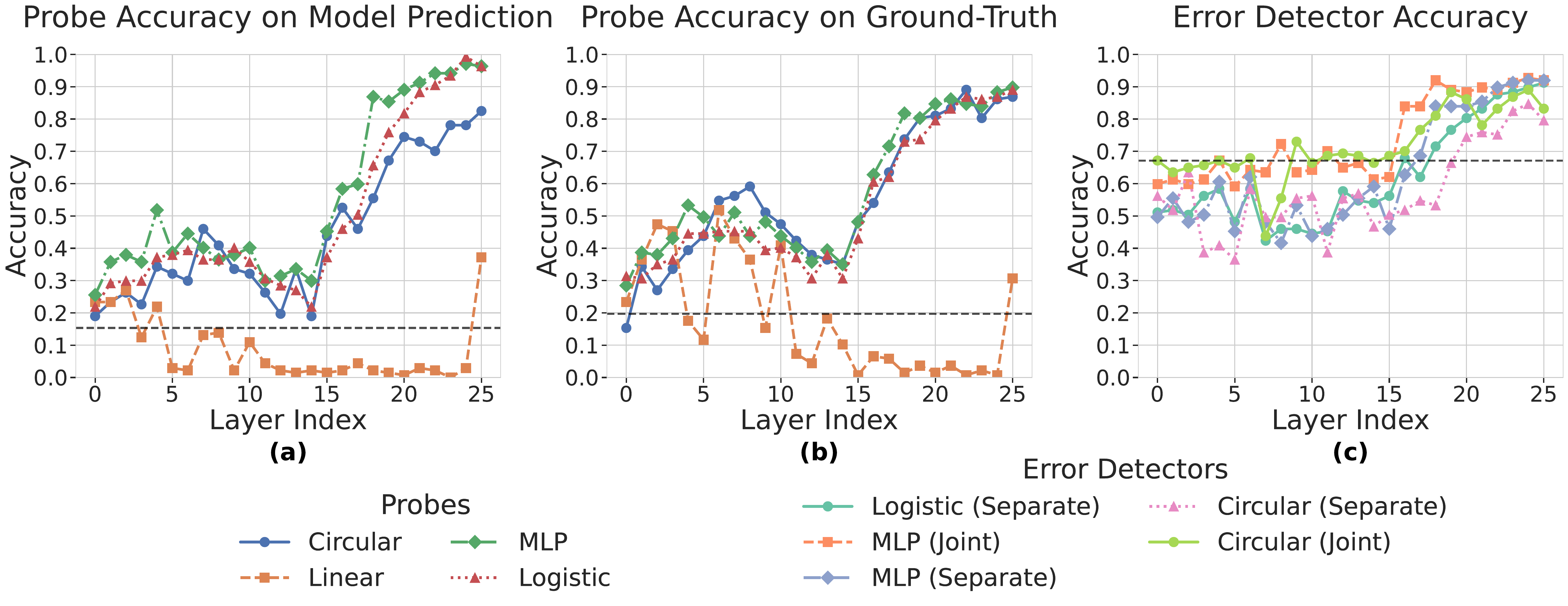}
    \caption{\textbf{Probing Structured Chain-of-Thought Reasoning.}
(a) Probes recover the model’s predicted digit with increasing accuracy across layers; non-linear probes reach over 90\% in the final layers.
(b) Ground-truth digits remain decodable, though slightly less accurately than in the pure arithmetic setting, likely due to added linguistic context.
(c) Error detectors achieve 80-90\% accuracy in later layers. The dashed lines indicate the accuracy of the majority class baseline.}
    \label{fig:gsm8k_setting_probes}
\end{figure*}

\subsection{Can Probes Predict Model Correctness?}
\label{sec:pure_correctness}

Having established that both the model’s predicted output and the ground-truth answer can be decoded from its internal representations, we now ask whether this information can be combined to predict when the model is likely to make a mistake. In other words, can probes anticipate whether the model’s answer will be correct or incorrect?

To test this, we adapt our probing setup to a binary classification task: predicting whether the first digit of the model’s output matches the ground-truth result. We experiment with five probing-based error detectors:
\begin{enumerate}
    \item \textbf{Separate Circular Probes.} Two circular probes are trained independently--one to predict the model’s output (as in Section 4.1), and one to predict the ground-truth answer (Section 4.2). The error predictor simply returns 1 if the two predictions disagree.
	\item \textbf{Joint Circular Probe.} Two circular probes are trained jointly. The angular difference between their raw predictions is passed through a sigmoid function, and the resulting scalar is trained with binary cross-entropy to predict correctness.
	\item \textbf{Separate MLP Probes.} Similar to the circular setup above, but using two independently trained MLP probes.
	\item \textbf{Single MLP Classifier.} A single MLP is trained directly to predict correctness as a binary classification task.
	\item \textbf{Separate Logistic Probes.} Two logistic probes are trained independently, and their predictions are compared via the same disagreement rule.
\end{enumerate}
All probes are trained on the same data used in prior experiments, with the only change being the training label: whether the first digit of the model’s output is correct. Additional implementation details for each method are provided in \cref{app:error_detection_probes}.

The results are shown in \cref{fig:pure_setting_probes}c. As expected, all detectors perform near chance level in early layers but improve substantially in the later layers of the model. Most approaches reach accuracies above 90\%, with some approaching 95\%, far above the 50\% majority baseline of the balanced dataset.
Overall, these results demonstrate that probes can effectively detect model errors based solely on internal representations. The fact that high-accuracy error detection can be achieved using simple probe architectures reinforces the finding that LLMs encode useful information about both their predictions and the correct answers, even when the two diverge.

\section{Probing Structured Chain-of-Thought Reasoning}
\label{sec:probe_gsm8k}
So far, we have shown that simple probes can effectively recover both model predictions and ground-truth results in a direct arithmetic setting, where the language model is prompted with isolated addition queries. We now move to a more challenging and practically relevant scenario: arithmetic reasoning embedded within chain-of-thought (CoT) traces \cite{wei2022chain}.
Specifically, we focus on addition-only problems from the GSM8K dataset \cite{cobbe2021trainingverifierssolvemath}, prompting Gemma 2 2B IT to solve them using a structured CoT intermediate reasoning, with each intermediate step formatted as \texttt{<a+b=c>}.

To construct a suitable dataset, we follow the abstraction approach used in prior work \cite{patel-etal-2021-nlp, stolfo-etal-2023-causal, opedal2024do, mirzadeh2025gsmsymbolic}. We first filter out problems that involve operations beyond addition. Then we augment this set of problems by abstracting away surface-level text by converting each question into a template form where numerical values are replaced by symbolic placeholders (e.g., \texttt{x}$_1$, \texttt{x}$_2$, ...), using Claude 3.7 Sonnet.\footnote{\url{https://www.anthropic.com/claude/sonnet}} We then generate concrete problem instances by sampling new operand values and substituting them into the templates.%
We prompt the model to format each intermediate computation step as \texttt{<a+b=c>}.
For each problem, we store the model’s full reasoning trace and extract a set of intermediate computation steps (i.e., the input and all prior steps up to a given index). The final dataset contains 685 problem steps, roughly balanced in terms of model correctness and hundred-digit. We divide it into a 80/20\% train-test split. Full dataset construction details are in \cref{app:appendix_gsm8k_data}.

As in the previous experiments, we train probes on the residual stream activations at the equals-sign token (\texttt{=}) of each CoT step. We replicate our previous structure: first probing for the model’s internal prediction (\cref{sec:gsm_output}), then the ground-truth result (\cref{sec:gsm_ground_truth}), and predicting model correctness (\cref{sec:gsm_correctness}). We then evaluate whether probes trained on simple 3-digit queries generalize to the CoT setting (\cref{sec:cross_setting}).

\subsection{Can Probes Predict the Model’s Output?}
\label{sec:gsm_output}

We begin by testing whether the model’s internal representations at each CoT step encode the output it is about to produce. Accuracy results are shown in \cref{fig:gsm8k_setting_probes}a.
As in the direct arithmetic setting, we observe that the linear probe performs poorly, while the MLP, logistic, and circular probes achieve strong performance in deeper layers. In this CoT setting, accuracy improves sharply between layers 15 and 20, with the MLP and logistic probes reaching approximately 92\% accuracy at the top layers. The circular probe performs slightly worse, plateauing at around 85\%.

\subsection{Can Probes Recover the Ground-Truth Answer?}
\label{sec:gsm_ground_truth}

We now turn to the task of predicting the ground-truth result of each intermediate step in the CoT reasoning chain. As before, we train probes to recover the correct answer from the residual stream activation at the equals-sign token. Results are shown in \cref{fig:gsm8k_setting_probes}b.
Consistent with our earlier findings, probe accuracy increases with model depth for all probe types except the linear one. The strongest probes (MLP, logistic, and circular) plateau in the final 3-4 layers, indicating that the model progressively builds more explicit representations of the correct answer over its depth.

However, compared to the pure arithmetic setting (\cref{sec:pure_ground_truth}), we observe a modest drop in overall accuracy, with performance ranging between 80\% and 90\% in the best layers. We attribute this to the increased complexity of the CoT setting: the language model’s input includes not only arithmetic expressions, but also natural language context describing the math word problem. This additional contextual information likely introduces noise that affects the clarity with which numerical information is encoded in the hidden states, making it slightly harder for the probes to recover the exact result.

\subsection{Can Probes Predict Model Correctness?}
\label{sec:gsm_correctness}

Finally, we evaluate whether probes can predict whether the model’s arithmetic predictions at intermediate CoT steps are correct. We adopt the same set of error detection strategies described in \cref{sec:pure_correctness}, applying them to the structured CoT setting.

The results are shown in \cref{fig:gsm8k_setting_probes}c. Although slightly noisier than in the pure arithmetic setting, the overall trend remains consistent. All error detectors perform at or near chance level until approximately layer 15, after which accuracy increases sharply and stabilizes between 80\% and 90\% in the final layers.
These findings indicate that the ability of simple probes to detect arithmetic errors is not limited to isolated arithmetic queries. Instead, this capability transfers well to more realistic, structured settings involving chain-of-thought reasoning.

\begin{figure}[t]
    \centering
    \includegraphics[width=0.33\textwidth]{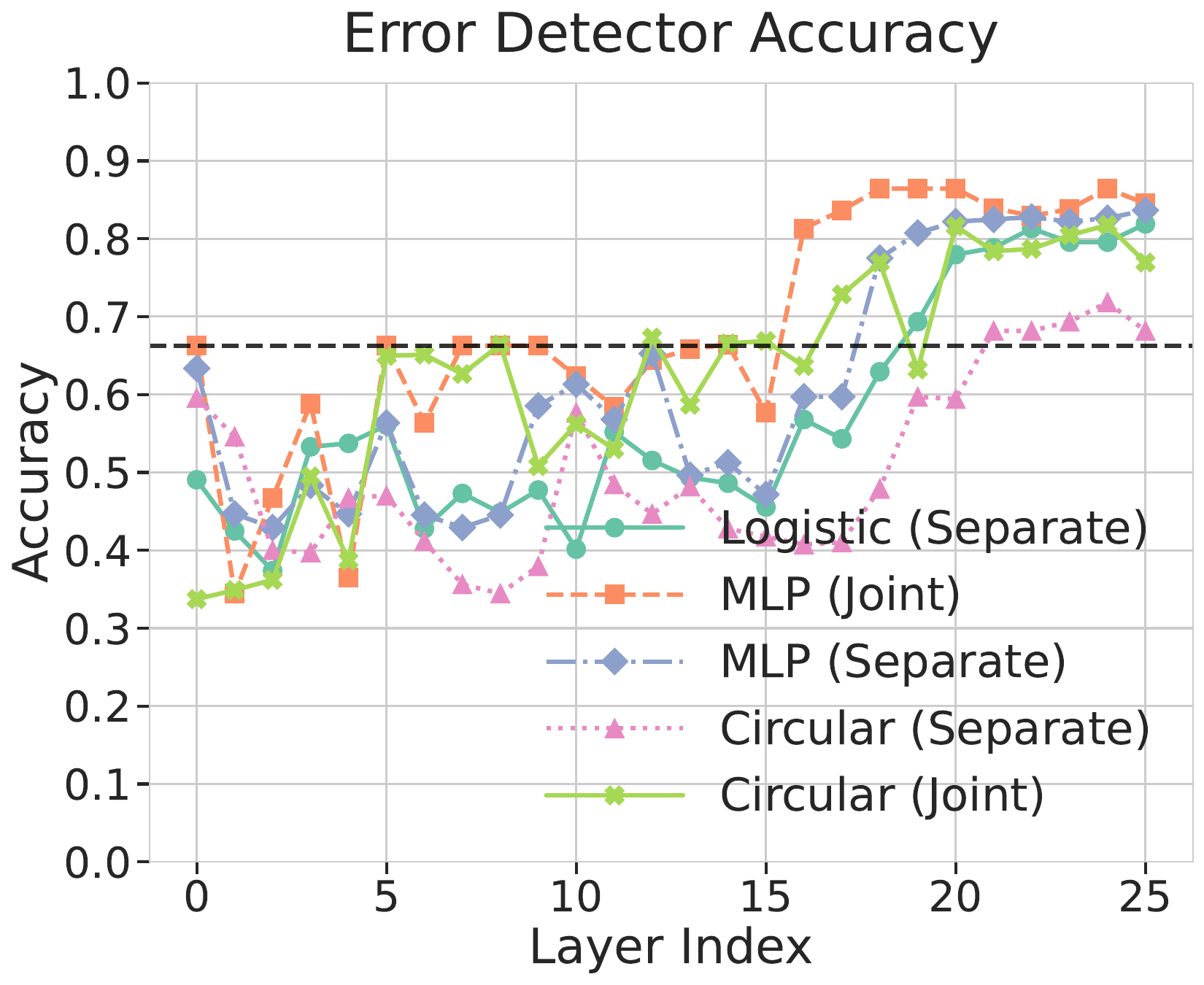}
    \caption{\textbf{Cross-Setting Error Detection.} Accuracy of probes trained on simple arithmetic queries evaluated on GSM8K problems.
Probes generalize well to the structured CoT setting, reaching up to 85\% accuracy and indicating consistent internal representations.}
    \label{fig:cross_setting_error_detector}
\end{figure}

\subsection{Do Probes Generalize Across Settings?}
\label{sec:cross_setting}
A natural question is whether probes trained in one setting (e.g., direct arithmetic queries) can generalize to another (e.g., arithmetic embedded in chain-of-thought reasoning). If the internal representations of numerical quantities are consistent across contexts, we would expect probes trained in one domain to retain predictive power when applied to the other.
To evaluate this, we test the error detectors trained in the pure arithmetic setting on the structured CoT dataset, without any further training. Results are reported in \cref{fig:cross_setting_error_detector}.

Remarkably, we find that these probes achieve nearly the same accuracy as those trained directly on the CoT data, with performance around 85\%. This suggests that the language model encodes numerical information in a robust and consistent manner, independent of whether the arithmetic appears in isolation or within a reasoning chain.
These findings point toward the potential for reusable probes that generalize across settings, offering a lightweight way to monitor model behavior in multi-step reasoning tasks.

\section{Using Error Detectors for Self-Correction}
\label{sec:application}

So far, we have shown that simple probes can reliably identify when a language model is likely to make an arithmetic error. We now explore a practical application of these probes: using them to guide the model in revising erroneous reasoning steps during multi-step arithmetic problems in the GSM8K dataset.

All error detectors used in this section are trained solely on data from the pure arithmetic setting, with supervision targeting only the hundreds digit of the result. As shown in \cref{sec:cross_setting}, these detectors generalize well to reasoning steps in GSM8K.

For all experiments below, we use the best-performing configuration: a single MLP-based error detector applied to the residual stream at layer 25. When a reasoning step is flagged as likely incorrect, we append a follow-up prompt immediately after the model’s output. This prompt takes the form of a brief corrective message (e.g., ``That step looks suspicious. Let’s re-do just this step:''), followed by the equation from the flagged step up to the equals sign (e.g., \texttt{<123+456=}). The message prompts the model to recompute the step without altering the rest of the chain. We experiment with several versions of the corrective message, ranging from neutral (e.g., “that step looks wrong”) to stronger wording (e.g., “that’s definitely wrong”). The full list of prompts is provided in Appendix E.

We evaluate this setup on a set of 685 intermediate steps from GSM8K, sampled to include both correct and incorrect predictions. Using full-number correctness as the evaluation criterion, the error detector flagged 178 true positives and 22 false positives.
\cref{tab:self_correction_updated} reports the results across different prompting styles. We evaluate two metrics: (i) TP Correction, the proportion of model errors (true positives) that are corrected after re-prompting, and (ii) FP Preservation, the proportion of correct steps (false positives) that remain unchanged after re-prompting. We find that simple feedback prompts can correct up to 11.8\% of model errors. At the same time, most prompts preserve all false positives, with a 100\% preservation rate in most configurations.

These results demonstrate that error detectors trained on simple arithmetic data can serve as effective weak oracles for self-correction, enabling language models to revise specific reasoning steps with minimal risk of degrading correct ones.

\begin{table}[t]\small
\centering
\begin{tabular}{lcc}
\toprule
\textbf{Message} & \textbf{TP Correction} & \textbf{FP Preservation} \\
\midrule
suspicious & 11.80\% & 100.00\% \\
neutral    & 11.80\% & 100.00\% \\
specific   & 10.11\% & 100.00\% \\
stronger   & 8.99\% & 95.45\% \\
detailed   & 6.18\%  & 100.00\% \\
\bottomrule
\end{tabular}
\caption{\textbf{Self-Correction via Re-Prompting.}
Different re-prompting messages lead to varying correction rates, with up to 11.8\% of flagged errors corrected and near-perfect preservation of correct answers.}
\label{tab:self_correction_updated}
\end{table}

\section{Related Work}
\paragraph{Arithmetic Capabilities in Language Models.}
Although several studies have highlighted the brittleness and inconsistency of language models’ reasoning abilities \cite{razeghi-etal-2022-impact, stolfo-etal-2023-causal, srivastava2024functionalbenchmarksrobustevaluation, hong2024evaluatingllmsmathematicalcoding, opedal2025mathgap, mirzadeh2025gsmsymbolic}, recent advancements have shown significant improvements in their performance on mathematical tasks \cite{lewkowycz2022solving, azerbayev2024llemma, nature, shao2024deepseekmathpushinglimitsmathematical}. These gains have motivated a line of work aimed at enabling models to self-reflect or self-correct their reasoning \cite{weng-etal-2023-large, madaan2023selfrefine, kim2023language, shinn2023reflexion, zhou2024solving, pan-etal-2024-automatically}. However, subsequent analysis has revealed that the improvements reported by some of these methods were partially due to the use of oracle labels during the correction process, rather than being fully model-internal \cite{huang2024large}. Our work aligns with these findings, demonstrating that even a weak oracle--here implemented as a simple probe--can effectively support model correction and improve arithmetic reasoning performance.

\paragraph{Interpretability of Arithmetic Reasoning.}
Several studies have investigated how language models encode and manipulate numerical information, both in small-scale transformers trained on synthetic tasks \cite{nanda2023progress, zhong2023the, quirke2024understanding,  ding2024survival, MALTONI2024106550}, and in pre-trained language models \cite{hanna2023how, stolfo-etal-2023-mechanistic, zhang2024interpreting,nikankin2025arithmetic, lindsey2025biology}. A particular area of focus has been the structure of numerical representations in language models. Recent work has shown that such quantities may be encoded in circular, linear, or helical geometries \cite{levy-geva-2025-language, kantamneni2025language, zhu-etal-2025-language}, connecting to broader studies of non-linear representations in transformers \cite{yedida-2, yedida-1, csordas-etal-2024-recurrent, shai2024transformers,engels2025not}.

\section{Conclusion}

We analyzed how language models encode arithmetic information and showed that simple probes can extract both predicted outputs and correct answers from internal activations. These probes can also anticipate errors, generalize to chain-of-thought reasoning, and serve as weak oracles for self-correction. When used in a re-prompting setup, they improve accuracy with minimal risk of degrading correct outputs.
These results are consistent with prior work showing that language models often encode the correct answer internally even when their output is incorrect \cite{orgad2025llms, gekhman2025insideouthiddenfactualknowledge}, and show that lightweight probing tools offer a practical path toward extracting and leveraging this latent knowledge.

\section*{Limitations}

While our findings provide evidence that simple probes can detect arithmetic errors from internal activations, several limitations remain.

First, our analysis focuses exclusively on addition problems.
In \cref{app:subtraction}, we extend the setup to include subtraction and observe consistent results. However,
it remains an open question whether the same probing strategies are effective for other arithmetic operations such as multiplication, or division. Extending the methodology to these operators would help assess the robustness and generality of our approach.

Second, our experiments are primarily conducted on a single model, Gemma 2B IT. We additionally report consistent results on Phi-3 in \cref{app:phi3}, but a broader evaluation across diverse architectures and scales would strengthen the generality of our conclusions.

Third, our chain-of-thought (CoT) analysis is limited to addition-only problems from the GSM8K dataset. While GSM8K provides a useful benchmark for arithmetic reasoning, future work should explore other datasets, including those with broader math coverage and more varied reasoning formats.

Finally, our CoT evaluation is conducted under a constrained setting where intermediate reasoning steps are explicitly formatted as structured equations (e.g., \texttt{<a+b=c>}). This structure simplifies probing and error detection. However, real-world model outputs are typically unstructured and expressed in natural language. An important direction for future work is to extend our approach to operate over general, unstructured CoT traces, with the goal of building lightweight tools that detect and correct reasoning errors in free-form outputs. Preliminary evidence in support of this extension is provided in \cref{app:free-form}.

\section*{Acknowledgments}
We would like to express our gratitude to Vilém Zouhar for useful discussions and comments on our work. AS acknowledges the support of armasuisse Science and Technology through a CYD Doctoral Fellowship.

\bibliography{custom}

\appendix

\clearpage

\section{Pure Arithmetic Data Generation}
\label{app:pure_data}

To train and evaluate digit-level probes and error detectors in the pure arithmetic setting, we generate synthetic data using a two-shot prompting setup with an instruction-tuned language model.
The prompting setup includes:

\paragraph{System Message:}
\begin{quote}
\texttt{You are a helpful assistant that calculates the sum of two numbers. Always provide your answer in the format <<x+y=z>> where x is the first number, y is the second number, and z is their sum. Do not provide any additional explanation.}
\end{quote}

\paragraph{Few-shot Examples:}
Each prompt includes two demonstrations, where the user asks:
\begin{quote}
\texttt{Calculate the sum of the following two numbers: \\
First number: \{i\} \\
Second number: \{j\}}
\end{quote}

The expected assistant response is:
\begin{quote}
\texttt{<<i+j=z>>}
\end{quote}

A new target question in the same format is then added to create a complete prompt. This setup is used to generate three-digit addition problems. We extract the residual stream activation at the equals sign (\texttt{=}) in the model's output to form input representations for probe training, with supervision targeting a specific digit (e.g., the hundreds digit) of the result.

We use a 2-shot prompt in the GSM8K setting mainly to ensure that the model adheres to the expected output format. To maintain consistency, we presented the results obtained with 2-shot prompting in the pure arithmetic setting as well. However, we additionally replicated our probing experiments for Gemma2-2b-it under 0-shot and 1-shot settings, and observed similarly strong probe performance across all configurations. Specifically, the best-performing error detectors achieved accuracies of 94.6\% (0-shot), 95.8\% (1-shot), and 95.4\% (2-shot). 

In the pure arithmetic setting, the accuracy of Gemma-2-2b-it under 0-shot, 1-shot, and 2-shot prompting is 98.3\%, 98.5\%, and 98.6\%, respectively. 

\section{Sampling Strategy}
We construct the probing dataset from model outputs on all addition problems $(a, b)$ where $a, b \in [100, 999]$ and $a + b < 1000$.

Samples are first filtered to include a mixture of both correct and incorrect model predictions. Specifically, the incorrect samples are those where the model's output differs from the true sum in the hundreds digit. 

All samples are then grouped based on the hundreds digit of the model's predicted output. From each digit class (2-9), we randomly select up to 100 samples, ensuring that both correct and incorrect predictions are represented across digit classes. If a digit class contains fewer than 100 samples, all available samples from that class are included. For Gemma 2 2B IT, this procedure yields exactly 800 examples. The final dataset is then split into training (70\%) and testing (30\%) sets.

This sampling strategy achieves:
\begin{itemize}
  \item A balanced distribution over digit classes (in terms of predicted hundreds digits);
  \item A representative mix of correct and incorrect predictions, supporting both value probing and error detection tasks.
\end{itemize}

\section{Error Detection Probes}
\label{app:error_detection_probes}
Here we provide the formulation for the error-detecting probes discussed in \cref{sec:pure_correctness,sec:gsm_correctness}.

\paragraph{MLP Error Detector (Single).}
The MLP consists of a single hidden layer with ReLU activation and a hidden dimensionality of 512. As in the logistic probe, the output layer produces 2 logits corresponding to correct and incorrect predictions. The prediction about model correctness $\hat{y}_\mathrm{c}$ is given by:
\begin{equation}\nonumber
    \hat{y}_\mathrm{c} = \arg\max_i \bigg(\mathbf{W}_2^\top \ \mathrm{ReLU}\Big(\mathbf{W}_1^\top \mathbf{x}_l + \mathbf{b}_1 \Big) + \mathbf{b}_2\bigg),
\end{equation}
where $\mathbf{W}_1 \in \mathbb{R}^{d_\mathrm{model} \times 512}, \mathbf{W}_2 \in \mathbb{R}^{512 \times 2}, \mathbf{b}_1 \in \mathbb{R}^{512}$, and $\mathbf{b}_2 \in \mathbb{R}^{2}$ are the probe's parameters. This probe  is trained using \textbf{cross-entropy loss} on binary labels (1 = correct, 0 = incorrect), and optimized using Adam.

\paragraph{Circular Error Detector (Joint).}
We combine two circular probes by taking angular difference between their outputs and passing it through a sigmoid to get an output probability value.
More formally, let the probe weights be two vectors $\mathbf{w}_1^{(1)}, \mathbf{w}_2^{(1)}, \mathbf{w}_1^{(2)}, \mathbf{w}_2^{(2)} \in \mathbb{R}^{d_\mathrm{model}}$. We compute the probe's output $\hat{y}$ as
\begin{align}
    \theta_1 =& \mathrm{atan2} \big(\mathbf{w}_1^{(1)\top}\mathbf{x}_l, \mathbf{w}_2^{(1)\top}\mathbf{x}_l \big),\\
    \theta_2 =& \mathrm{atan2} \big(\mathbf{w}_1^{(2)\top}\mathbf{x}_l, \mathbf{w}_2^{(2)\top}\mathbf{x}_l \big),\\
    \hat{y}_\mathrm{c} =&\ \sigma \big(\theta_1 - \theta_2 \big).
\end{align}
The probe is trained with binary cross-entropy loss and the weights for the two circular probes are optimized jointly.

\paragraph{Separate Training Strategy.}
In this approach, two separate digit probes (e.g., logistic, MLP, or circular) are trained independently: one probe is trained to predict the model's actual output ($\hat{y}_\mathrm{model}$), while the other is trained to predict the ground truth label ($\hat{y}_\mathrm{GT}$).
At inference, error detection is performed by checking whether the two probes agree:
\begin{equation}
    \hat{y}_\mathrm{c} = \mathbbm{1}\big[ \hat{y}_\mathrm{model} = \hat{y}_\mathrm{GT}  \big]
\end{equation}
This procedure does not require any supervision and can be applied post hoc to existing probes.

\section{GSM8K Dataset Construction for Controlled Reasoning}
\label{app:appendix_gsm8k_data}

To enable consistent probing and self-correction experiments on symbolic arithmetic within chain-of-thought (CoT) reasoning, we construct a controlled subset of GSM8K problems involving only addition.

\paragraph{Step 1: Filtering addition-only problems.}  
We first identify GSM8K questions whose solutions involve only addition operations. This ensures that intermediate steps are structurally aligned with the arithmetic patterns studied in the pure setting.

\paragraph{Step 2: Abstracting numerical structure.}  
We use Claude 3.7 Sonnet in \textit{extended thinking} mode to extract and abstract each problem’s numerical content into symbolic variables (e.g., \texttt{x}$_1$, \texttt{x}$_2$, ..., \texttt{x}$_n$), creating reusable symbolic templates.
We manually verified 20 examples produced by this pipeline and found the variable abstraction to be accurate in all cases.

\paragraph{Step 3: Sampling concrete values.}  
We sample all variables independently and uniformly from the range $[100,999]$. This process allows for consistent arithmetic structure while introducing variation across generated problems.

\paragraph{Example instance.}
One resulting example from our dataset is:

\begin{quote}
\textbf{Question:} Sarah is planning to do some baking. She buys 771 pounds of rye flour, 611 pounds of whole-wheat bread flour, and 505 pounds of chickpea flour. Sarah already had 758 pounds of whole-wheat pastry flour at home. How many pounds of flour does she now have? \\
\textbf{Symbolic variables:} $x_1=771$, $x_2=611$, $x_3=505$, $x_4=758$ \\
\textbf{Expected reasoning:}
\begin{align*}
\texttt{<<} 771+611  &= 1382\texttt{>>} \\
\texttt{<<}1382 + 505 &=  1887\texttt{>>} \\
\texttt{<<}1887 + 758 &=  2645\texttt{>>}
\end{align*}
\textbf{Model response:}
\begin{align*}
\texttt{<<} 771 + 611 + 505  &= 1987\texttt{>>} \\
\texttt{<<}1987 + 758 &=  2745\texttt{>>} 
\end{align*}
\textbf{Predicted answer:} 2745 (incorrect)
\textbf{Correct answer:} 2645
\end{quote}

\paragraph{Model Performance of Gemma2-2b-it}
In the structured CoT setting, 83\% of intermediate steps are correctly formatted (i.e., of the form <<a+b=c>>), and among these, 94.1\% yield the correct result. The overall per-problem accuracy, accounting for all outputs including formatting errors, is 29.6\%.

\begin{figure*}[t]
    \centering
    \includegraphics[width=0.92\textwidth]{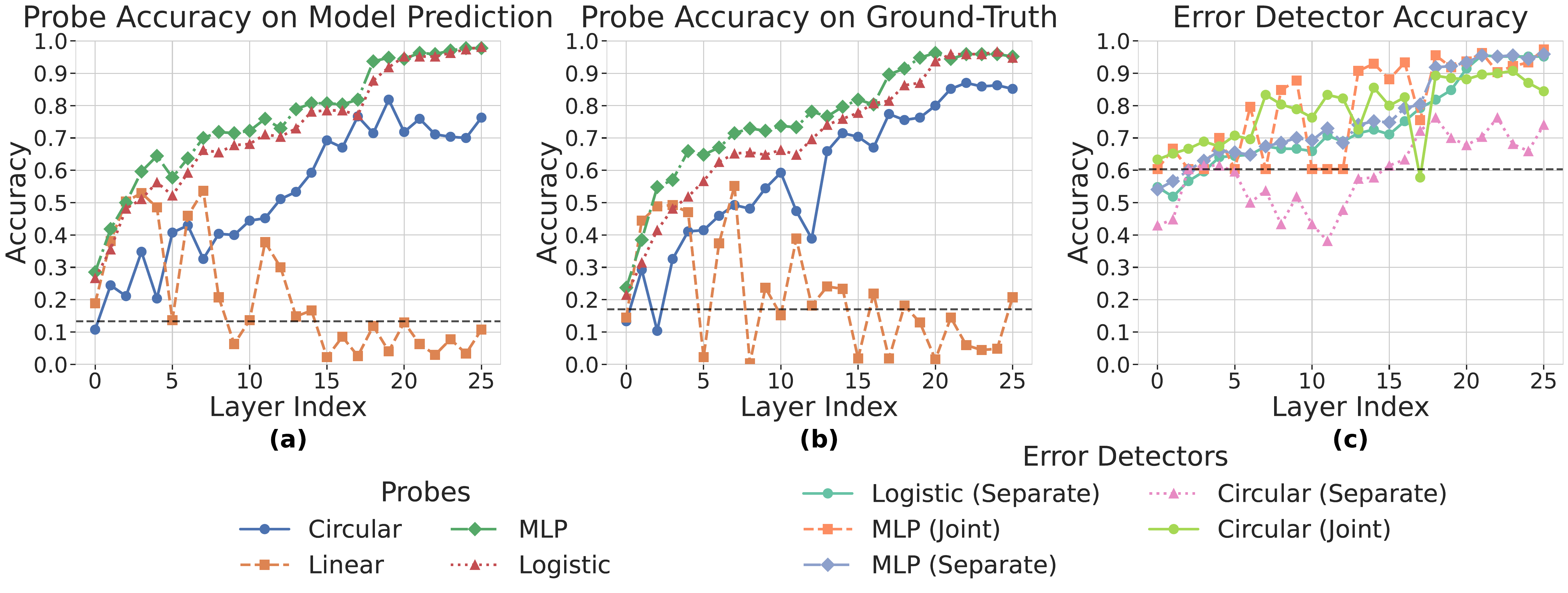}
    \caption{\textbf{Probing 3-Digit Arithmetic Queries (Subtraction).} Probing and error detection results on subtraction tasks. As with addition, probes reliably decode both model predictions (a) and ground-truth answers (b), and error detectors achieve high accuracy in predicting correctness (c), confirming the robustness of our findings.}
    \label{fig:gemma_difference}
\end{figure*}

\section{Prompting and Sample Selection in GSM8K}
\label{sec:appendix_gsm8k_prompting}

For the self-correction experiments described in Section~\ref{sec:application}, we use a series of structured prompts to intervene when the error detector flags a potentially incorrect step.

\paragraph{Prompt format.}  
Each prompt includes a short corrective message, followed by the previous equation up to the equals sign (\texttt{<<a + b =}). The following messages are used:

\begin{itemize}
    \item \textbf{suspicious:} \textit{"That step looks suspicious. Let's re-do just this step:"}
    \item \textbf{neutral:} \textit{"That step looks incorrect. Let's re-do just this step:"}
    \item \textbf{specific:} \textit{"The calculation in this step is incorrect. Let's recalculate:"}
    \item \textbf{stronger:} \textit{"That's definitely wrong. The correct calculation should be:"}
    \item \textbf{detailed:} \textit{"I made an error in adding these numbers. Let me compute the sum correctly step by step:"}
\end{itemize}

\paragraph{Sample selection.}  
In both the probing and intervention experiments on GSM8K, we select a single step per response for evaluation:

\begin{itemize}
    \item We first generate the full model response to a given question.
    \item We identify all equations in the format \texttt{<<a + b = c>>} and determine whether they are correct.

    \item If all equations are correct, we select the first one and extract the residual stream activation at the equals sign (\texttt{=}) as a \textbf{correct} sample.
    \item If any equation is incorrect, we select the first incorrect one and extract the activation at its equals sign as an \textbf{error} sample.
\end{itemize}

This means we test at most one equation per model response. The same procedure is used both for probe training and evaluation, and for the prompting-based self-correction experiments.

\section{Error Statistics}
\label{app:error_analysis}
We evaluated Gemma 2 2B IT on 319{,}972 samples of the form $(x, y)$ where $x + y < 1000$. The model produced 3,329 (1.04\%) incorrect results. Among these errors: 537 ($\sim$16\% of the total errors) had incorrect ones digits,  2,203 ($\sim$66\%) had incorrect tens digits. 3,328 ($\sim$100\%) had incorrect hundreds digits.

\begin{figure*}[t]
    \centering
    \includegraphics[width=0.92\textwidth]{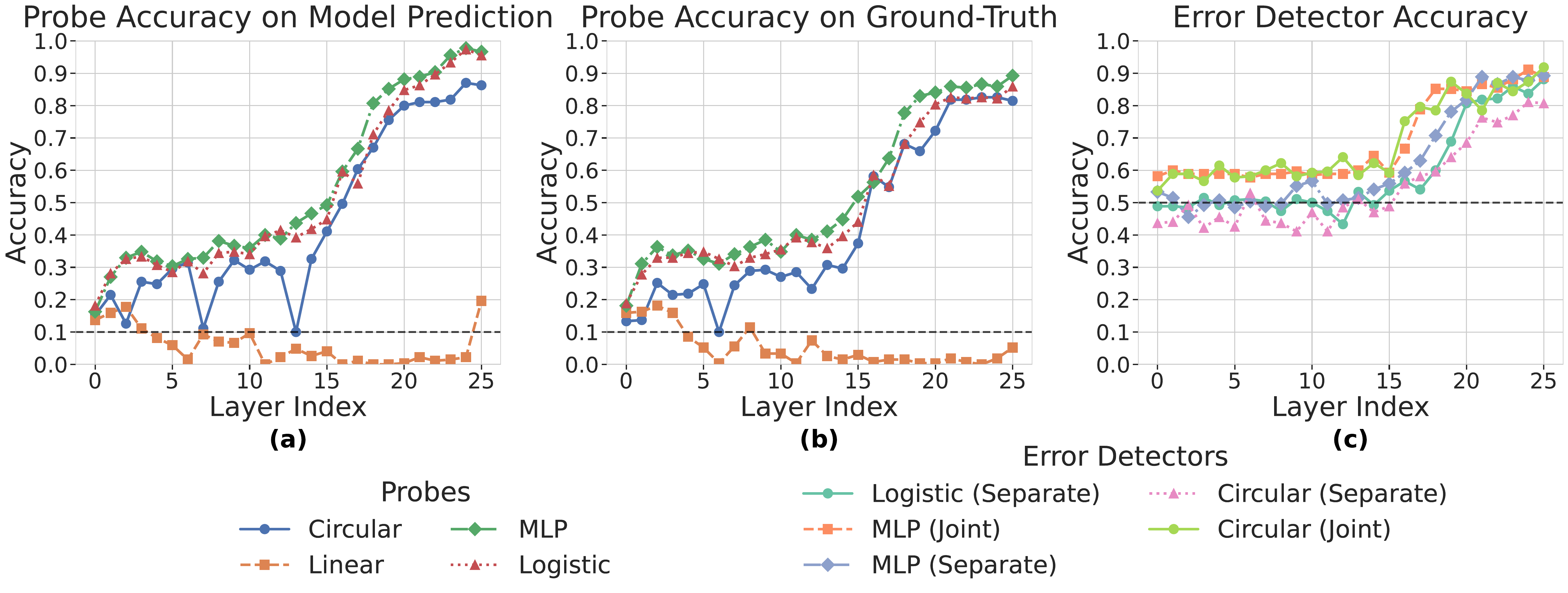}
    \caption{\textbf{Probing 3-Digit Arithmetic Queries (Wider Operand Range).} Probing and error detection results on an wider operand range. As with addition, probes reliably decode both model predictions (a) and ground-truth answers (b), and error detectors achieve high accuracy in predicting correctness (c), confirming the robustness of our findings.}
    \label{fig:gemma_3to5}
\end{figure*}

\section{Computational Resources and Tools}
\label{app:exp_details}
All experiments were conducted on a single Nvidia RTX 4090 GPU. The longest training run for the probes took 1 hour. Generating model response for the experiments in \cref{sec:probe_gsm8k} took 90 GPU hours. %
Our experiments were carried out using \texttt{PyTorch} \cite{NEURIPS2019_bdbca288} and HuggingFace \texttt{transformers} \cite{wolf-etal-2020-transformers}. All models were run with \texttt{bfloat16} precision for efficient memory usage.
For some of our experiments, we used a subset of the GSM8K dataset \cite{cobbe2021trainingverifierssolvemath}, which is available through an MIT license.\footnote{\url{https://choosealicense.com/licenses/mit/}}
We performed our data analysis using \texttt{NumPy} \citep{harris2020array} and \texttt{Pandas} \citep{mckinney-proc-scipy-2010}. The paper's bibliography was curated using \texttt{Ryanize-bib} \cite{ryanizebib}.

\section{Additional Results}
\label{app:additional_results}
In this section, we extend our main findings by evaluating the generality of our probing methodology across a broader range of settings. In particular, we consider arithmetic tasks involving both addition and subtraction (\cref{app:subtraction}), problems with a wider operand range (up to 5-digit numbers; \cref{app:wider_operand_range}), and free-form CoT reasoning with unconstrained natural language outputs (\cref{app:free-form}).

\subsection{Generalization to Subtraction}
\label{app:subtraction}
We construct a dataset with 800 addition and 800 subtraction queries, randomly mixing them. We train and evaluate digit-level probes and error detectors following the procedure in~\cref{sec:probe_arith}. Results in~\cref{fig:gemma_difference} show that the probes reach accuracy levels comparable to the addition-only case, indicating that they remain effective even when multiple arithmetic operators are present.

\subsection{Generalization to Wider Operand Range}
\label{app:wider_operand_range}
We expand the operand range in the pure addition setting to include numbers from 100 to 99,999. We generate a total of 900 examples, ensuring balance across first-digit values. As shown in~\cref{fig:gemma_3to5}, probes trained on this data continue to achieve strong performance on predicting both model outputs and ground-truth results, as well as detecting correctness, showing robustness to operand scale.

\subsection{Generalization to Free-Form CoT}
\label{app:free-form}
We let Gemma-2 2B IT generate responses to addition-only questions from our augmented version of GSM8K in a free format. Instead of enforcing a structured format (e.g., \texttt{<a+b=c>}), we use a traditional “let’s think step by step” prompt and encourage the model to produce a bullet-pointed list of intermediate reasoning steps (e.g., ``\texttt{1. [step 1]\textbackslash n 2. [step 2]}''), where each step may contain a mix of natural language and numerical computations. To obtain evaluation data for our digit-level probes and error detectors, we use GPT-4.1-mini\footnote{\url{https://openai.com/index/gpt-4-1}} to extract the operands and the model’s predicted result at each reasoning step. We manually verify a randomly selected subset of 20 examples to confirm that the extraction is accurate and reliable.

We then evaluate the probes trained in the 3-to-5 digit ``pure addition'' setting (\cref{app:wider_operand_range}) on this free-form CoT data. For each reasoning step, we locate the token immediately preceding the first digit of the model’s prediction and extract the residual stream activation at that position for probing. The evaluation set is balanced such that a majority class baseline would yield 50\% accuracy. Despite the increased variability of this free-form setting, our probes and error detectors maintain non-trivial performance (as shown in \cref{table:free_cot_probe,table:free_cot_error}). In particular, the best-performing error detector achieves 72\% accuracy, demonstrating that our methodology generalizes beyond highly structured setups.

\begin{table}[t]
\centering
\begin{tabular}{lcc}
\toprule
\textbf{Probe Type} & \textbf{GT Accuracy} & \textbf{Output Acc.} \\
\midrule
Linear   & 0.0454 & 0.0306 \\
MLP      & 0.6901 & 0.8547 \\
Circular & 0.4926 & 0.6629 \\
Logistic & 0.6413 & 0.8695 \\
\bottomrule
\end{tabular}
\caption{\textbf{Performance of Different Probe Types on Free-Form CoT Data.} We report the accuracy in predicting both the ground-truth digit and the model’s output digit at each reasoning step. Results correspond to the maximum accuracy achieved by each probe type across all layers.}
\label{table:free_cot_probe}
\end{table}

\begin{table}[t]
\centering
\begin{tabular}{lcccc}
\toprule
\textbf{Error Detector Type} & \textbf{Accuracy} \\
\midrule
Logistic Separately & 0.6754 \\
MLP                 & 0.7208 \\
MLP Separately      & 0.7026 \\
Circular Separately & 0.6129  \\
Circular Jointly    & 0.6901  \\
\bottomrule
\end{tabular}
\caption{\textbf{Accuracy of Different Error Detector Configurations Applied to Free-Form CoT Data.} Despite the increased variability in reasoning steps, detectors trained on structured arithmetic still generalize well. Results correspond to the maximum accuracy achieved by each probe type across all layers.}
\label{table:free_cot_error}
\end{table}

\begin{figure*}[t]
    \centering
    \includegraphics[width=0.92\textwidth]{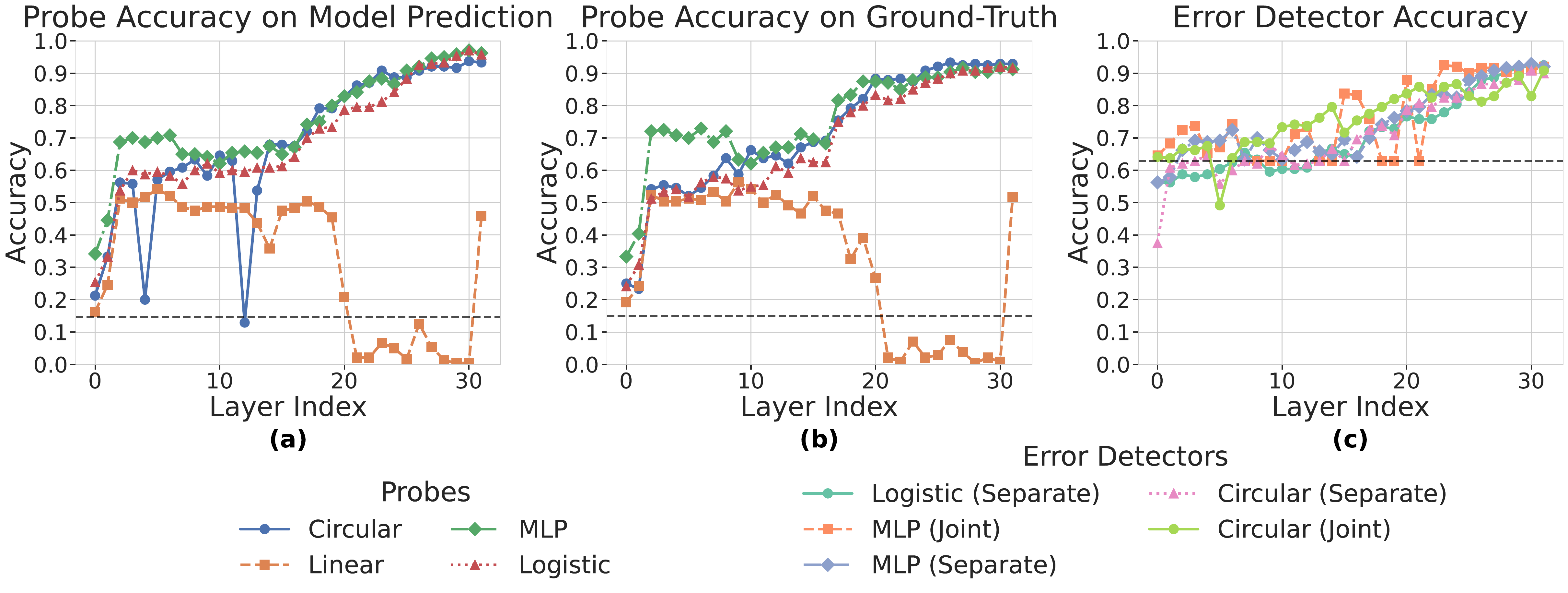}
    \caption{\textbf{Probing 3-Digit Arithmetic Queries (Phi-3).} Probing and error detection results on Phi-3. As with Gemma 2 2B IT, probes reliably decode both model predictions (a) and ground-truth answers (b), and error detectors achieve high accuracy in predicting correctness (c), confirming the robustness of our findings across models.}
    \label{fig:phi_pure}
\end{figure*}

\begin{figure*}[t]
    \centering
    \includegraphics[width=0.92\textwidth]{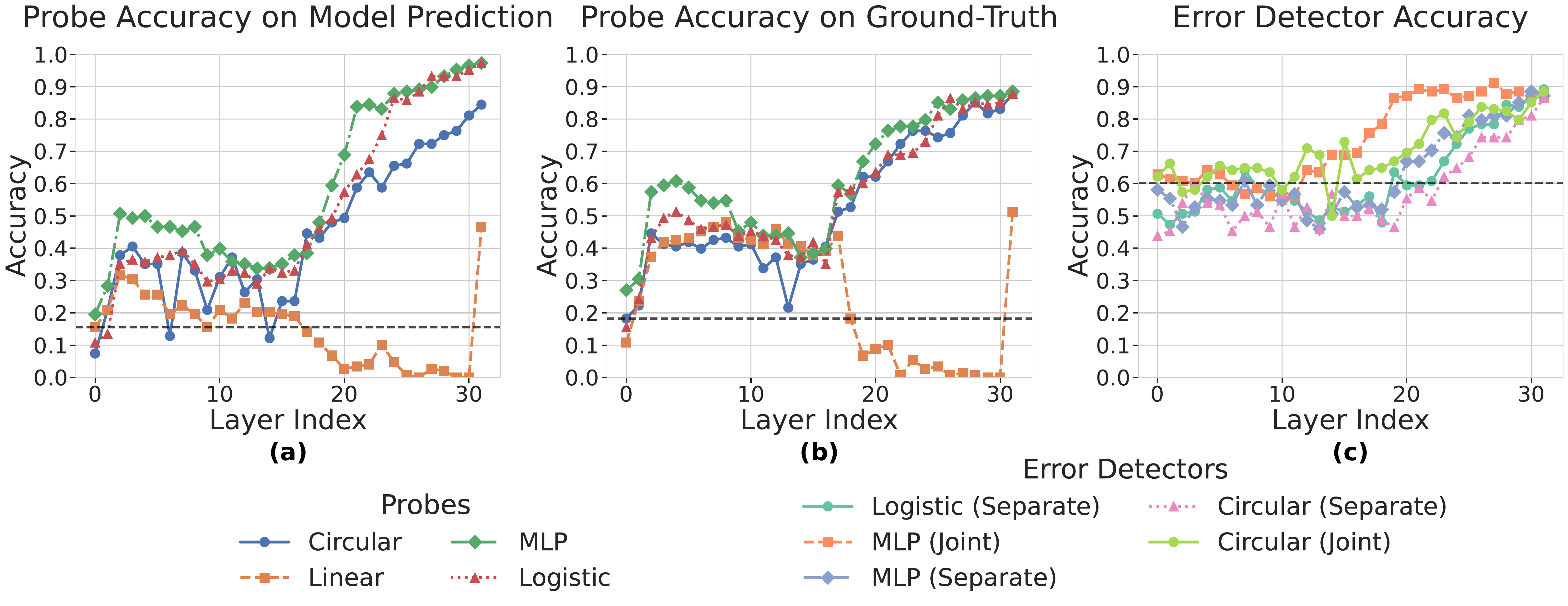}
    \caption{\textbf{Probing Structured Chain-of-Thought Reasoning (Phi-3).} (a) Probes accurately recover the model’s prediction in deeper layers. (b) Ground-truth digits are similarly decodable. (c) Error detectors achieve strong performance, confirming that findings generalize to structured CoT reasoning with Phi-3.}
    \label{fig:phi_gsm}
\end{figure*}

\section{Generalization to Phi-3}
\label{app:phi3}

To test the generality of our findings, we replicate our probing and error detection experiments on Phi-3 \cite{abdin2024phi3technicalreporthighly} a 3.8B parameter language model with architecture and training data distinct from Gemma 2B IT.\footnote{HuggingFace ID: \texttt{microsoft/Phi-3-mini-4k-instruct}.}

We follow the same prompting and evaluation procedures as in \cref{sec:probing_intro,sec:probe_gsm8k}. Probes and error detectors are trained using residual stream activations at the equals-sign token in both the pure arithmetic and GSM8K settings.

Overall, we observe results on Phi-3 that are consistent with those observed on Gemma 2 2B IT. In the pure arithmetic setting (\cref{fig:phi_pure}), probes successfully recover both the model’s predicted digit and the ground-truth result, with MLP, logistic, and circular probes reaching over 90\% accuracy in the final layers. In addition, error detectors trained on these signals achieve high accuracy in predicting model correctness, surpassing 90\% in deeper layers. Overall, these findings reinforce the generality of our main results. Despite differences in architecture, training, and tokenization, both models exhibit similar representational trends.

The same takeaways apply also for the structured CoT setting (\cref{fig:phi_gsm}), where probes recover both the model’s prediction and the ground-truth answer with high accuracy in deeper layers, and error detectors reliably identify incorrect steps. %

\section{Ethical Considerations}

This work explores how simple probing techniques can be used to detect arithmetic errors from the internal activations of language models. While primarily intended to improve model reliability and transparency, this capability may introduce potential risks.

One concern is that probing tools could be used to reverse-engineer or extract sensitive information from model internals in settings where the model has been fine-tuned on proprietary or confidential data. Although our experiments are limited to synthetic arithmetic tasks, similar techniques might be adapted to more sensitive domains.

Another consideration is the use of probing-based feedback mechanisms to automatically modify or steer model behavior. While we focus on error correction in a controlled setting, improperly validated corrective feedback could introduce bias or reinforce incorrect reasoning patterns in more open-ended tasks.

Lastly, while our methods are lightweight and interpretable, they might be incorrectly interpreted as offering guarantees of correctness or safety. We emphasize that probes are statistical tools, and any system incorporating them should be evaluated rigorously before deployment in high-stakes settings.

\end{document}